\pdfoutput=1

\documentclass[11pt]{article}
\usepackage{amsmath,amsfonts}
\usepackage{bbm}
\usepackage{algorithmic}
\usepackage{algorithm}
\usepackage{array}
\usepackage{textcomp}
\usepackage{stfloats}
\usepackage{url}
\usepackage{verbatim}
\usepackage{graphicx}
\usepackage{hyperref}
\usepackage{subfigure}
\usepackage{cite}
\usepackage{multirow}
\usepackage{stfloats}
\usepackage[final]{acl}

\usepackage{times}
\usepackage{latexsym}

\usepackage[T1]{fontenc}

\usepackage[utf8]{inputenc}

\usepackage{microtype}

\usepackage{inconsolata}

\usepackage{graphicx}
\usepackage{color}

\title{Pseudo-Label Enhanced Prototypical Contrastive Learning \\ for Uniformed Intent Discovery}

\author{
    Yimin Deng\textsuperscript{1,2}$^{*}$,
    Yuxia Wu\textsuperscript{3}$^{*}$,
    Guoshuai Zhao\textsuperscript{1}$^{\dagger}$,
    Li Zhu\textsuperscript{1}$^{\dagger}$,  
    Xueming Qian\textsuperscript{4}
    \\
    \textsuperscript{1}School of Software Engineering, Xi'an Jiaotong University \\
    \textsuperscript{2}City University of Hong Kong,
    \textsuperscript{3}Singapore Management University\\
    \textsuperscript{4}School of Information and Communication Engineering, Xi'an Jiaotong University
    \\
    \text{dymanne@stu.xjtu.edu.cn, yieshah2017@gmail.com}\\
 \text{\{guoshuai.zhao,zhuli\}@xjtu.edu.cn, qianxm@mail.xjtu.edu.cn }
  } 

\begin{document}
\maketitle

\begingroup
\renewcommand\thefootnote{\relax}
\footnotetext{$^{*}$ Co-first authors with equal contribution.}
\footnotetext{$^{\dagger}$ Corresponding authors.}
\endgroup

\begin{abstract}


New intent discovery is a crucial capability for task-oriented dialogue systems. Existing methods focus on transferring in-domain (IND) prior knowledge to out-of-domain (OOD) data through pre-training and clustering stages. They either handle the two processes in a pipeline manner, which exhibits a gap between intent representation and clustering process or use typical contrastive clustering that overlooks the potential supervised signals from the whole data. Besides, they often individually deal with open intent discovery or OOD settings. To this end, we propose a \textbf{P}seudo-\textbf{L}abel enhanced \textbf{P}rototypical \textbf{C}ontrastive \textbf{L}earning (\textbf{PLPCL}) model for uniformed intent discovery. We iteratively utilize pseudo-labels to explore potential positive/negative samples for contrastive learning and bridge the gap between representation and clustering. To enable better knowledge transfer, we design a prototype learning method integrating the supervised and pseudo signals from IND and OOD samples. In addition, our method has been proven effective in two different settings of discovering new intents. Experiments on three benchmark datasets and two task settings demonstrate the effectiveness of our approach.\footnote{The codes and datasets are available at https://github.com/dymanne123/PLPCL }
\end{abstract}
\section{Introduction}
\begin{figure}[t]
    \centering
    \centering
   \includegraphics[height=0.22\textwidth]{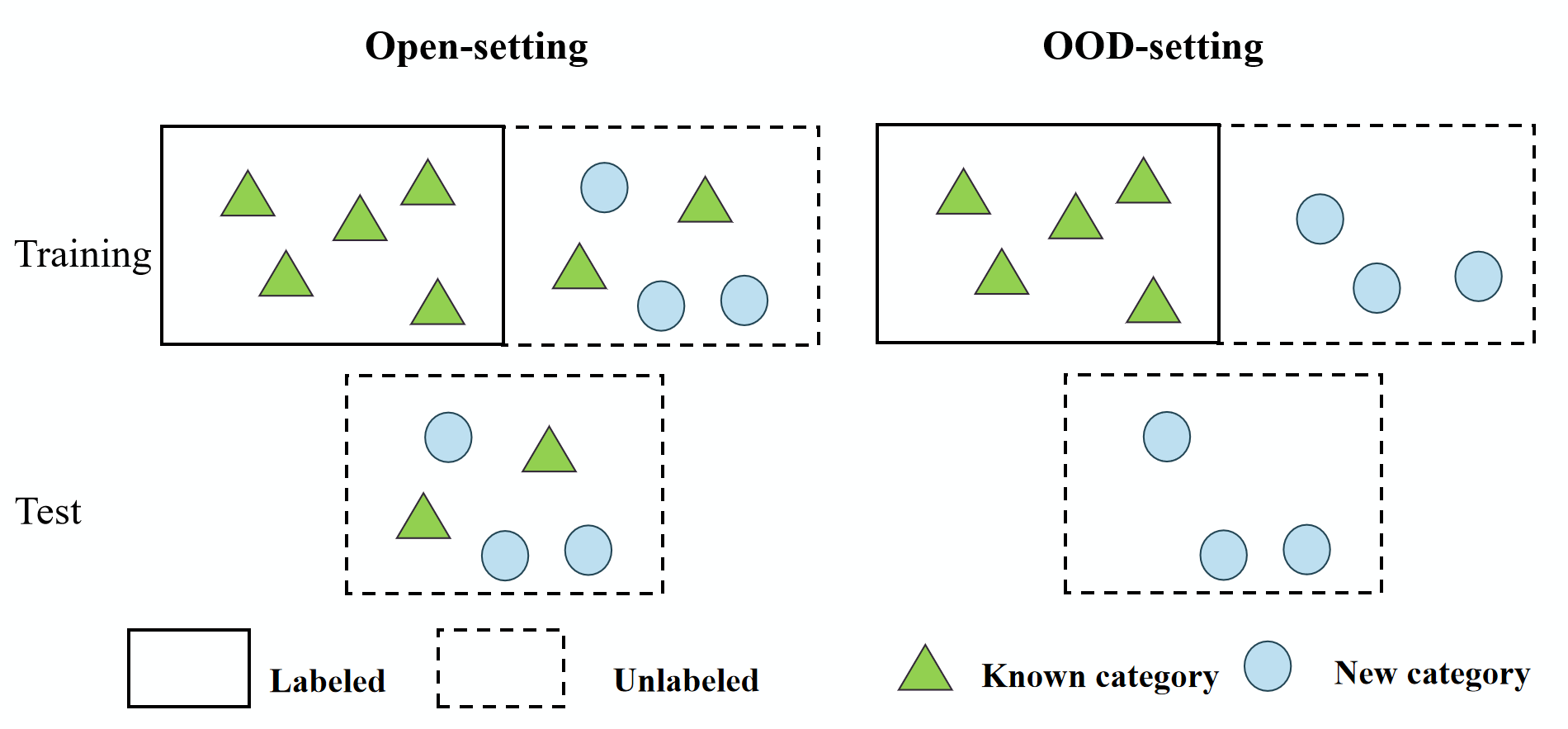}
    \caption{Two basic task settings for uniformed intent discovery. 
    Open-setting: Partially labeled IND data is used for training, and the test data includes both IND and OOD categories. OOD-setting: Fully labeled IND data is used for training, while the test data contains only OOD categories.
    }
    \label{setting}
\end{figure}

Task-oriented conversational systems are typically designed to assist users in completing specific tasks by leveraging predefined ontologies (\textit{e.g.,} intents, slots and values), which face limitations in dynamic environments where novel ontology categories may emerge \cite{zhang2021discovering, mou2022watch, wu2022state,wu2024active,liang2024synergizing}.
New intent discovery, which aims to uncover and categorize out-of-domain intents absent from the training data, has received increasing attention due to its crucial role in dialogue systems \cite{vedula2019towards,min2020dialogue}. Initially, researchers focused on exploring unsupervised clustering methods \cite{hakkani2015clustering,shi2018auto,liu2021open}. However, real-world scenarios often involve limited labeled data, prompting a shift toward semi-supervised approaches, notably OOD and open intent discovery \cite{mou2022disentangled,mou2022watch,lin2020discovering,zhang2021discovering,shen2021semi,zhang2022new,liang2023clusterprompt, liang2024actively}. OOD intent discovery involves clustering unlabeled OOD intents by utilizing labeled IND data, focusing solely on identifying novel categories absent in the training set. In contrast, open intent discovery seeks to simultaneously recognize both known and new categories from unlabeled data, allowing for a more comprehensive understanding of both in-domain and out-of-domain intents (see Figure \ref{setting}).

The existing approaches for new intent discovery typically use a two-stage contrastive clustering approach, involving IND pre-training and OOD clustering. For open intent discovery, researchers focus on effectively utilizing the small amount of labeled data for weakly supervised \cite{lin2020discovering} or semi-supervised clustering methods \cite{zhang2021discovering,shen2021semi}. For OOD intent discovery, previous works commonly employ a contrastive clustering framework \cite{li2021contrastive} with approaches such as multi-head contrast learning \cite{mou2022disentangled} or neighbor-enhanced contrastive strategies \cite{mou2022watch}. 

Although previous methods in intent discovery have achieved notable success, several challenges in the field remain unexplored. 
(1) One key challenge is efficiently integrating labeled and unlabeled data for joint representation learning and clustering. While some methods focus on contrastive clustering for joint learning \cite{mou2022disentangled}, they often overlook critical supervised signals from IND samples during the clustering stage. Leveraging labeled information for this process remains underexplored, particularly given the limited availability of labeled data and the risk of overfitting. (2) Another critical issue is devising effective transfer learning mechanisms between IND and OOD data while preventing catastrophic forgetting. Existing methods often discard classifiers trained on prior knowledge, retaining only feature extraction during clustering with OOD samples \cite{zhang2021discovering}. This requires additional alignment strategies, potentially introducing noise if suboptimal. There's an urgent need for methods preserving prior knowledge while adapting better to new intent data, ensuring seamless transfer learning. (3) Moreover, the predominant focus of prior research has been either on open intent discovery or OOD intent discovery individually, disregarding the practical need for a unified method capable of handling both scenarios \cite{zhang2021discovering, mou2022disentangled}. Real-world dialogue systems often encounter situations requiring updates or migration, underscoring the need for a uniform intent discovery approach that can adapt to various system changes. 

To address these limitations, we propose a \textbf{P}seudo-\textbf{L}abel enhanced \textbf{P}rotypical \textbf{C}ontrastive \textbf{L}earning (PLPCL) model which is built upon contrastive clustering for joint representation learning and clustering. Our approach begins by pre-training the contrastive clustering model using labeled IND data. To effectively harness labeled information, we integrate IND samples with unlabeled data using a semi-supervised clustering strategy, employing distinct contrastive learning strategies for labeled and unlabeled data. To prevent overfitting and maximize the utilization of unlabeled data, we iteratively select reliable unlabeled samples with confident pseudo-labels. These reliable samples serve as potential positive/negative samples during contrastive learning, enhancing the overall contrastive clustering process.

To bridge the gap between IND and OOD data, we introduce a prototype learning strategy. It maintains the prototype matrix by integrating instance and cluster features from both the IND and reliable unlabeled samples. Our method integrates contrastive clustering and prototypical learning, eliminating the need for an extra aligning module. This integration facilitates improved knowledge transfer from IND to OOD without discarding label information from IND samples. Furthermore, our framework is purposefully designed for uniform intent discovery, demonstrating effectiveness in both open-setting and OOD-setting scenarios.

The contribution of our work is threefold: (1) We introduce a novel method that leverages both labeled and unlabeled data through pseudo-label enhanced semi-supervised contrastive learning. This approach facilitates joint representation learning and clustering by effectively leveraging the whole data. (2) We propose a prototypical contrastive learning model for uniformed intent discovery integrating prototypical learning and contrastive clustering to bridge the gap between IND prior knowledge and OOD categories. (3) We conduct extensive experiments on both OOD and open intent discovery scenarios and the results demonstrate the effectiveness of our proposed method.

\section{Related Work}

\begin{table*}[t]
\centering
\small
\resizebox{1\textwidth}{!}{
\begin{tabular}{l|c|c|c|c|c}
\hline
                & Joint representation and cluster   & Prior knowledge retention   & Reuse of labeled data&OOD setting&Open setting    \\ \hline
CDAC+ \cite{lin2020discovering} &  \checkmark&	pair-wise similarity&\checkmark&\texttimes&	\checkmark
 \\
DeepAligned \cite{zhang2021discovering}        &\texttimes	&representation&\texttimes&\texttimes	&\checkmark

 \\ 

DKT \cite{mou2022disentangled}        & \checkmark	&representation&\texttimes&\checkmark	&\texttimes

 \\
 DPN \cite{an2023generalized} & \texttimes&prototype&\checkmark&\texttimes&\checkmark\\
 PLPCL (Ours) &\checkmark&representation, prototype, classifier &\checkmark&\checkmark&\checkmark\\
 \hline
\end{tabular}}
\caption{The differences between our method and prior works}
\label{tab:novelty}
\end{table*}
\subsection{Intent Discovery} 
Recent research for intent discovery can be broadly categorized into OOD-setting and open-setting.
As shown in Figure \ref{setting}, open intent discovery involves clustering both IND and OOD intents with IND priori knowledge. Samples with IND intents are not all labeled. Whereas OOD intent discovery focuses on accurately handling OOD intents and assumes that the intents of labeled and unlabeled data do not overlap, which means all IND samples are labeled. 
\citet{lin2020discovering} proposed a self-supervised clustering method that utilizes limited labeled data. \citet{zhang2021discovering} proposed a k-means-based semi-supervised clustering method that can effectively use prior knowledge in intent discovery. \citet{mou2022disentangled} proposed a disentangled contrastive learning framework that mainly focuses on OOD intent clustering and decouples instance and cluster-level features to disentangle the knowledge of IND and OOD samples. 
 \citet{han2019learning} extended deep embedded clustering to transfer learning setting, incorporating prior knowledge for OOD clustering.


\subsection{Contrastive Clustering}
Contrastive clustering has been widely used in various clustering scenarios, such as unsupervised semantic segmentation \cite{hamilton2022unsupervised} and generalized self-supervised contrastive learning \cite{hu2022your}. It has also been applied in OOD intent detection and discovery tasks \cite{kumar2022intent,mou2022disentangled,mou2022watch}. \citet{li2021contrastive} proposed a contrastive clustering framework with two contrastive learning heads.
It provided objective guidance for clustering, avoiding interference from prior knowledge. \citet{mou2022disentangled} extended contrast clustering to the semi-supervised scenario and designed a two-stage contrastive learning process that includes both supervised pre-training and unsupervised clustering. It achieved state-of-the-art results for OOD intent discovery.



\subsection{Prototype Learning.}
Prototype learning methods are widely used in clustering analysis and classification problems. 
 In semi-supervised clustering scenarios, coarsely assigned pseudo-labels may result in mismatches between instances and prototypes, introducing noise that significantly affects the clustering performance \cite{wu2022semi,an2023generalized, huang2022learning}. \citet{an2023generalized} used weighted pseudo-labels to reduce the effect of mismatched prototypes. \citet{huang2022learning} proposed the approach of prototype scattering, which enhances the variance between the clusters by maximizing the distances between prototype features, to obtain well-separated clusters.
The prototype learning method is robust to noise and outliers. Compared to other clustering methods, it is an intuitive and interpretable approach that can provide references for the entire cluster based on representative examples.

\subsection{Novelty Analysis of Our Method}
We summarize the differences between our method and prior works and highlight the motivations and advantages of our method in terms of joint representation and clustering, prior knowledge retention, reuse of labeled data, and handling of OOD and open settings in Table \ref{tab:novelty}.

\section{Preliminaries}


\subsection{Problem Statement} 
\begin{figure*}[t]
    \centering
    \resizebox{1\linewidth}{!}{
    \includegraphics[height=0.15\textwidth]{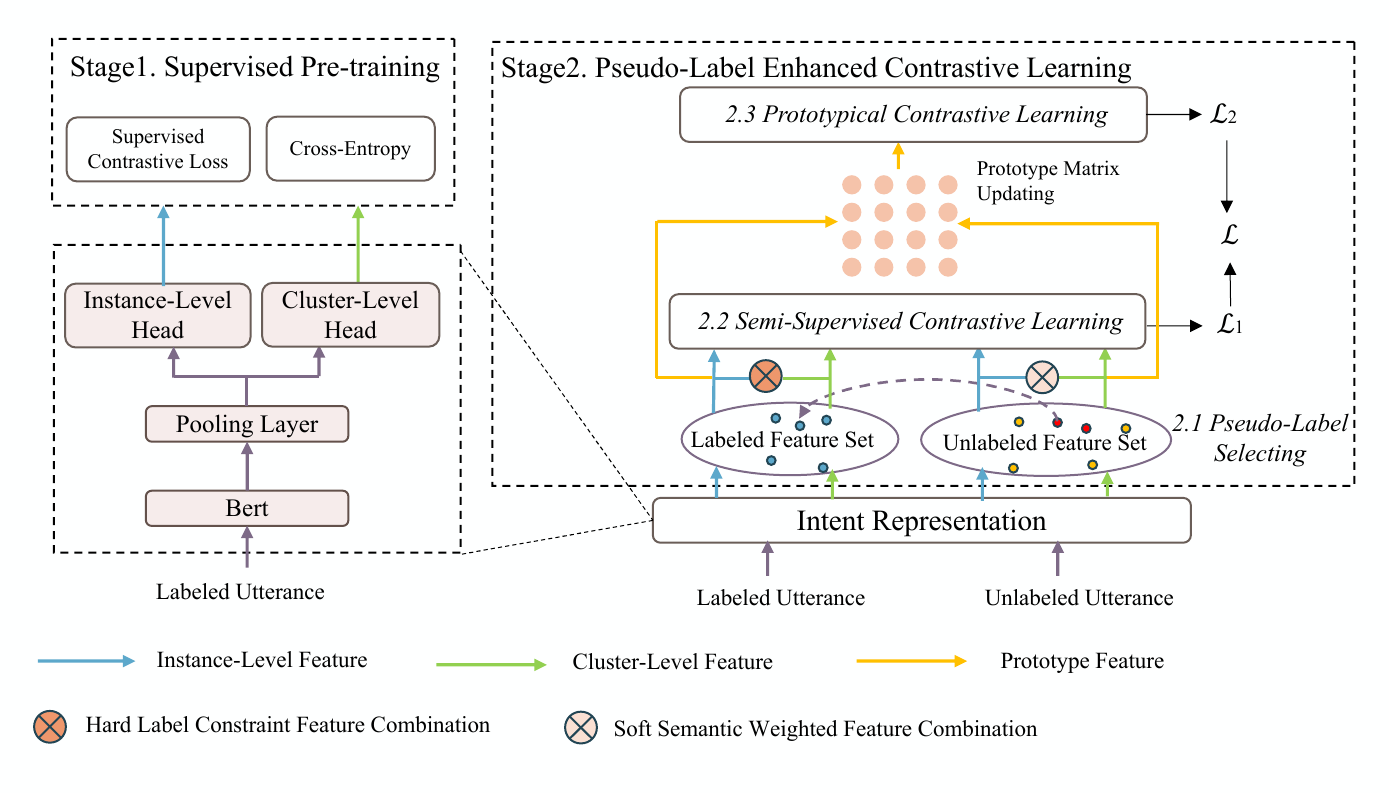}}
    \caption{The overall architecture of the proposed PLPCL. (a) Intent representation is achieved based on disentangled instance-level and cluster-level heads. (b) Stage 1 involves supervised contrastive learning on instance-level representation and classification on cluster-level representation. (c) Stage 2 starts from pseudo-label selecting for unlabeled data followed by semi-supervised and prototypical contrastive learning in an iterative manner.   
    }
    \label{model}
\end{figure*}
 Uniformed intent discovery includes both OOD intent discovery and open intent discovery. OOD intent discovery assumes we have a set of labeled IND data $D_{IND}$ and unlabeled OOD data $D_{OOD}$, with the goal of clustering OOD intents. Notably, there is no overlap between the IND and OOD data. 
 Open intent discovery involves an intent analysis dataset $\{D_l, D_u\}$, where $D_l=\{(x_l,y_l)|y_l\in \mathcal{Y}_k\}$ represents labeled data for known intents, and $D_u=\{(x_u,y_u)|y_u\in \mathcal{Y}_k, \mathcal{Y}_{uk}\}$ includes both known ($\mathcal{Y}_k$) and unknown intents ($\mathcal{Y}_{uk}$). In the extreme case, all samples of the known categories in the training set are labeled, and $D_u$ contains only data with unknown intents, making the prior labeling information same with OOD classification. Since we do not have a priori assumptions that $y_l \cap y_u=\emptyset$, this setting differs from the generalized intent discovery \cite{mou2022generalized}. 
 

\subsection{Contrastive Clustering} 

Our model is based on a contrastive clustering framework. It performs instance-level and cluster-level contrastive learning. Specifically, positive/negative instance pairs are formed via data augmentation, followed by their projection into a feature space. Instance-level and cluster-level contrastive learning are performed in the row space and column space, respectively.

Unsupervised instance-level contrastive learning (ILCL)  is performed on unlabeled data, where the augmented sample of each sample is considered as a positive sample and other samples are considered as negative samples. $f_i$,$f_j$ refer to the augmented samples that are generated from the same samples after passing through the dropout layer. 
\begin{equation}
\begin{split}
\mathcal{\ell }^{ins}_{i,j} = -log \frac{\exp \left(sim(f_{i} , f_{j}) / \tau\right)}{\sum_{k=1}^{2N}1_{i \neq k} \exp \left(sim(f_{i}, f_{k}) / \tau\right)}
\end{split}
\label{ILCL}
\end{equation}
On the cluster-level contrastive learning head $g$, it performs cluster-level contrastive learning (CLCL). The cluster representation of the augment sample is considered as a positive sample, and the other cluster representations are considered as negative samples. $y_i$ refers to the representations of the clusters, which are columns in the cluster-level feature matrix. $y_j$ are the dropout-augment representations for the cluster level.
\begin{equation}
\begin{split}
\mathcal{\ell }^{clu}_{i,j} = -log \frac{\exp \left(sim(y_{i} , y_{j}) / \tau\right)}{\sum_{k=1}^{2N}1_{i \neq k} \exp \left(sim(y_{i}, y_{k}) / \tau\right)}
\end{split}
\label{CLCL}
\end{equation}

\section{Methods}
The overall framework of PLPCL is illustrated in Figure \ref{model}. Our framework follows a two-stage pipeline. In the pre-train stage, IND-labeled samples are utilized for supervised multi-head contrastive learning to acquire prior knowledge. This model is adapted from contrastive clustering and includes intent representation alongside two independent heads. These heads are instrumental in decoupling the representation into instance-level and cluster-level spaces, facilitating joint representation learning and clustering. 

After pretraining, the prototypes of known categories are obtained, serving as a foundation for efficient knowledge transfer across IND and OOD data. In the second stage, the multi-head contrastive model is further trained on the entire dataset including IND and OOD samples. Specifically, this stage comprises three iterative steps: pseudo-label selecting, semi-supervised contrastive learning, and prototypical contrastive learning. These steps collectively aim to transfer prior knowledge to new categories and enhance the model's adaptability.



\subsection{Intent Representation}

To facilitate effective knowledge transfer between IND and OOD samples, we aim to achieve joint intent representation and clustering by learning instance-level and cluster-level representations. 

Drawing inspiration from \cite{mou2022disentangled}, we first extract the intent representation using a pre-trained BERT model and a pooling layer to extract text representation. Then we utilize two independent MLPs to map the intent representation $z_i$ into two disentangled latent vectors: $f_i = f(z_i)$ and $g_i = g(z_i)$. 



\subsection{Supervised  Pre-training}

To familiarize the model with the prior knowledge obtained from labeled IND samples and to establish initial cluster prototypes, we conduct pre-training on IND samples.  Based on the multi-level intent representation, we conduct two-level pre-training.   

For instance-level representation, we adopt supervised contrastive learning (SCL) to maximize inter-class variance and minimize intra-class variance within the IND samples. 

Formally, for a sample $x_i$ in a mini-batch of size $N$, the samples within $N$ sharing the same label are considered as positive samples, while the remainder is treated as negative samples. The SCL loss is computed as follows:
\begin{equation}
\begin{split}
\mathcal{L}_{S C L} = &\sum_{i=1}^{N}-\frac{1}{|N_{y_i}-1|} 
    \sum_{j=1}^{N} \mathbbm{1}_{i \neq j}\mathbbm{1}_{y_i = y_j}\\
    &\log \frac{\exp \left(f_{i} \cdot f_{j} / \tau\right)}{\sum_{k=1}^{N}\mathbbm{1}_{i \neq k} \exp \left(f_{i}\cdot f_{k} / \tau\right)}
\end{split}
\label{SCL}
\end{equation}
where $y_i$, $y_j$ are the labels of samples $x_i$, $x_j$ and $\mathbbm{1}$ is an indicator function. $N_{y_i}$ denotes the number of samples in $N$ with the label $y_i$. $f_i$, $f_j$ indicate the instance-level representation of $x_i$, $x_j$. $\tau$ is the temperature parameter for contrastive learning.


For the cluster-level representation, we apply cross-entropy loss (CE) to learn cluster-friendly features. Note that we use a classifier with both IND and OOD classes for open-setting to better reserve the priori knowledge extracted at this stage, allowing our model to better retain the prior knowledge acquired during the pre-training stage compared to previous works.

\subsection{Semi-supervised Training}

After pretraining with labeled data, we achieve a good initialization of representation learning and clustering. The challenge now lies in transferring the prior knowledge to new intents. There are two critical problems to be addressed: (1) Effectively utilizing both labeled and unlabeled data to enhance the joint representation and clustering process; (2) Transferring learned knowledge from IND to OOD data while continually refining representation learning to enhance cluster-friendly features without encountering catastrophic forgetting. To tackle these challenges, we introduce a pseudo-label enhanced contrastive learning scheme tailored for iterative clustering and updating. This scheme starts from reliable pseudo-label filtering for unlabeled samples, followed by semi-supervised contrastive learning and prototypical contrastive learning. 

\subsubsection{Pseudo-label Selecting}
\label{method-pseudo}

It's important to note that the multi-level intent representation heads are pre-trained on limited labeled data. To fully leverage the valuable information embedded within unlabeled data, we employ pseudo-labeling techniques to iteratively select unlabeled samples as weak supervised signals for subsequent contrastive learning processes.

For semi-supervised contrastive learning (\ref{method-SCL}), reliable pseudo-labeled data are amalgamated with labeled data, augmenting potential contrastive samples. Regarding prototypical contrastive learning (\ref{method-PCL}), pseudo-labels are employed to enrich the learning process of the prototype matrix. Furthermore, the integration of pseudo-labeled data introduces supplementary constraints to mitigate overfitting and enhance the model's generalization performance to novel unseen categories.



However, it's important to note that the quality of pseudo-labels is crucial, as noisy or incorrect pseudo-labels can degrade model performance. It is crucial to ensure the accuracy and reliability of pseudo-labels to maintain the effectiveness of the model's training. We treat pseudo-labels with probabilities greater than a confidence threshold as true labels and use them as supervised signals to guide model training:
\begin{equation}
  p(k|\textbf{x})>\sigma
\label{filtering}
\end{equation}
where $p(k|\textbf{x})$ denotes the probability that x belongs to class k, $\sigma$ represents confidence probability. When a certain probability is greater than the confidence threshold, we have enough confidence to consider it as belonging to this category. The choice of confidence probability will impact the strength of the supervised signal and the introduced noise. A high confidence probability will result in inadequate supervised information, whereas a low confidence probability will introduce erroneous pseudo-labels. 

\begin{table*}[tp]
\centering
\small
\resizebox{0.85\textwidth}{!}{
\begin{tabular}{c c c c c c c}
\hline

\hline
             Dataset& Classes  & Classes-IND   & Classes-OOD    &Training &Validation&Test   \\ \hline
BANKING   & 77 & 54 & 23 &9003&1000&3080 \\ 
STACKOVERFLOW & 20 & 14 &6 &12000&2000&6000 \\
CLINC & 150 & 105 &45 &18000&2250&2250 \\
\hline

\hline
\end{tabular}}
\caption{ Statistics of BANKING, STACKOVERFLOW and CLINC datasets.}
\label{tab:dataset}
\end{table*}

\subsubsection{Semi-supervised Contrastive Learning}
\label{method-SCL}

We utilize distinct contrastive learning strategies for IND labeled data and OOD unlabeled data. We compare confidence probabilities of pseudo-labels with a predefined confidence threshold, and if they are greater than this threshold, we consider them as reliable pseudo-labels.
Unlabeled samples with reliable pseudo-labels are considered as labeled data. SCL is applied on the instance-level contrastive learning head $f$ for labeled data, while unsupervised instance-level contrastive learning (ILCL) is performed on unlabeled data.

On the cluster-level contrastive learning head $g$, we perform cross-entropy loss for labeled data and perform cluster-level contrastive learning (CLCL) for OOD classes (OOD-setting) or all classes (open-setting).
During the training process, the number of unlabeled samples with reliable pseudo-labels will gradually increase. 


\subsubsection{Prototypical Contrastive Learning}
\label{method-PCL}

Decoupling knowledge from different levels is beneficial for separating the features of the source domain and target domain, thereby improving the efficiency of transfer learning and reducing overfitting. However, 
previous work only disentangled the instance-level and cluster-level features and applied constraints on them independently, without considering the inherent connection between the two levels of features.  Each sample potentially belongs to a cluster, and each cluster consists of a certain number of samples. 
In order to extract the relationship between instance features and cluster features and enhance the discrimination between clusters, we propose to maintain a cluster prototype matrix, which is of size $k*m$ and stores the prototype features of each cluster. 

 For each batch, the output of $f$ is an $n*m$ matrix containing the feature vectors of each sample, \textit{n} is the batch size and m is the feature vector dimension. The output of $g$ is an $n*k$ matrix, with each row corresponding to the probability that a sentence belongs to each class, \textit{i.e.,} $p(k|x)$, and each column corresponds to the representation of a cluster. The cluster prototype matrix is computed by averaging the instance-level representations over all samples belonging to the class. For labeled data and unlabeled data with reliable pseudo-label, we use ground truth or pseudo-label, referred to as hard label constraint feature combination; for other unlabeled data, we perform probability-weighted calculations, known as soft semantic weighted feature combination. As shown in the equation, \textit{G} is the cluster-level feature matrix $[g_1;g_2;\cdots;g_m; 1_{y_{m+1}},1_{y_{m+2}},\cdots, 1_{y_N}]$, $F$ is the instance-level feature matrix $[f_1;f_2;\cdots;f_N]$, $m$ is the number of labeled data in this batch. $G'$ and $F'$ are the cluster-level and instance-level feature matrix of the augmented samples, respectively.  
\begin{equation}
  M_{c} = G^T F
\end{equation}

The cluster prototype matrix $M_c$ is a $k*m$ matrix consisting of the features of each clustering center and can be written as $[m_1, m_2, \cdots, m_K]$. The obtained vector $m_i$ is normalized and used as the clustering center $z_i$ as shown in the equation.
\begin{equation}
 z_{c,i}=\frac{m_i}{\|m_i\|_2}
\end{equation}

After explicitly decoupling the cluster prototype vector $z_i$, the augmented features of each cluster prototype are used as positive samples, and the rest of the features are used as negative samples for contrastive learning at the cluster prototype level, as shown in the equation. Optimization of prototype contrastive loss (PCL) enables pulling apart different clusters and thus enhancing the discrimination between categories.
\begin{equation}
\begin{split}
\mathcal{\ell }^{pcl}_{i,j} = -log \frac{\exp \left(sim(z_{c,i} , z_{c,j}) / \tau\right)}{\sum_{k=1}^{2N}1_{i \neq k} \exp \left(sim(z_{c,i}, z_{c,k}) / \tau\right)}
\end{split}
\label{PCL}
\end{equation}

The final loss in the training process is obtained by combining SCL, ILCL, CE, CLCL and PCL.\footnote{We simply set the weight coefficients of each loss to 1. We compared the effects of different weights for supervised losses in appendix \ref{sec:loss weights}.}

\section{Experiments}

\subsection{Datasets}
We conduct experiments on three public datasets BANKING \cite{casanueva2020efficient}, STACKOVERFLOW \cite{xu2015short}  and CLINC \cite{larson2019evaluation}. 
BANKING consists of 13,083 queries covering 77 intents in the banking domain, and STACKOVERFLOW dataset contains 20 intents related to the programming domain. CLINC is a cross-domain intent dataset covering 150 intents across 10 domains.
Detailed statistics are shown in Table \ref{tab:dataset}, the division of the training set, validation set and test set remains consistent with previous works.
We take 30\% categories as unknown categories in both datasets, and all data with known intent is labeled. 

\begin{table*}[t]
\centering

\resizebox{1\textwidth}{!}{%
\begin{tabular}{ll|lll|lll|lll}
\hline

\multicolumn{2}{c|}{\multirow{2}{*}{Method}}                                          & \multicolumn{3}{c|}{BANKING-OOD} & \multicolumn{3}{c|}{Stackoverflow-OOD} &\multicolumn{3}{c}{CLINC-OOD}  \\
                                                         &                & ACC      & ARI      & NMI      & ACC      & ARI      & NMI    & ACC  & ARI     & NMI       \\ \hline

\multicolumn{2}{l|}{DTC\_BERT \cite{hsu2017learning}}              & 45.76   & 42.88     & 69.12    & 57.83    & 32.31    & 37.29 &51.56    & 48.33    & 84.64     \\
                                                  \multicolumn{2}{l|}{KCL\_BERT \cite{han2019learning} }                      & 47.61     & 36.5    & 64.51   & 41.33    & 28.74    & 34.42 & 57.33    & 49.45    & 80.35     \\
                                                  \multicolumn{2}{l|}{MCL\_BERT \cite{hsu2019multi}}              & 45.87   & 34.85     & 62.83    & 42.39    & 27.04    & 33.71 & 51.7    & 43.77    & 77.69	       \\
                                                  \multicolumn{2}{l|}{CDAC+ \cite{lin2020discovering} }                            & 59.78    & 44.58    & 69.19 & 61.56    &28.22     &52.76 & 73.04    & 64.44    & 87.90         \\
                                                  \multicolumn{2}{l|}{DeepAligned \cite{zhang2021discovering}}                      &63.86     &\underline{52.84}     &\underline{73.66}      & 79.68    & 63.18  & 65.52& 91.56    & 86.58    & 94.91       \\
                                                  \multicolumn{2}{l|}{DSSCC \cite{kumar2022intent} }            &64.67 & 51.38&  71.25    &\underline{84.28}&  65.94&  64.75&80.89&  73.55&  89.40    \\
                                                  \multicolumn{2}{l|}{DKT \cite{mou2022disentangled} }            &66.50   &52.07      &72.22    &82.22	&61.53	&\textbf{67.05}&\underline{94.96}	&\underline{90.25}	&\underline{95.94}    \\
                                                  
                                                  \multicolumn{2}{l|}{DPN \cite{an2023generalized} }            &\underline{68.15}&	49.72	&\textbf{74.76}   &78.9	&66.49	&66.85&87.33	&82.94	&95.95 \\
                                                  \multicolumn{2}{l|}{Llama2 \cite{touvron2023llama}}              & 27.82   & 45.26     & 3.25    & 71.24    & \underline{67.62}    & 48.63 &26.37&	4.19	&56.98 \\
                                                    \hline
                                                  
                                                  \multicolumn{2}{l|}{PLPCL} & \textbf{68.37}    & \textbf{53.19}     & 72.04 &\textbf{86.28}&\textbf{69.64}&\underline{66.95}&\textbf{95.11}&\textbf{90.71}&\textbf{96.15}      
                                                       \\ \hline

\end{tabular}
}
\caption{The OOD-setting results on three datasets. Overall 1$^{st}$/2$^{nd}$ in \textbf{bold}/{\underline{underline}}. We randomly sample 30\% of all classes as OOD intents for both datasets. Results are averaged over three random runs. (p $<$ 0.05 under t-test) 
}
\label{tab:main_result1}
\vspace{-0.3cm}
\end{table*}

\begin{table*}[t]
\centering
\resizebox{1\textwidth}{!}{%
\begin{tabular}{ll|lll|lll|lll}
\hline

\multicolumn{2}{c|}{\multirow{2}{*}{Method}}                                          & \multicolumn{3}{c|}{BANKING-Open} &  \multicolumn{3}{c|}{Stackoverflow-Open}&\multicolumn{3}{c}{CLINC-Open} \\
                                                         &               & ACC  & ARI     & NMI& ACC  & ARI     & NMI& ACC  & ARI     & NMI     \\ \hline

\multicolumn{2}{l|}{DTC\_BERT \cite{hsu2017learning}}               & 42.56	&	31.72&69.12    &52.7     &35.19      & 49.3&50.22     &39.72      & 79.51    \\
                                                  \multicolumn{2}{l|}{KCL\_BERT \cite{han2019learning} }                      & 64.87	&	54.52&80.07    &63.43     &50.42      & 61.83&69.02     &62.98      & 88.77 \\
                                                  \multicolumn{2}{l|}{MCL\_BERT \cite{hsu2019multi}}              & 65.39	&	55.21&79.53    &63.55     &47.51      & 57.18 &68.4     &60.68      & 87.78   \\
                                                  \multicolumn{2}{l|}{CDAC+ \cite{lin2020discovering} }                                  & 45.00	&33.10	& 69.49   & 67.05    &48.66      & 66.03  & 51.64    &35.89      & 79.96    \\
                                                  \multicolumn{2}{l|}{DeepAligned \cite{zhang2021discovering}}                     & \underline{74.84}	&\underline{64.37}	&84.86    &    \underline{76.77} & \underline{59.42}     & 71.97 &    \textbf{88.57} & \textbf{83.71}     & \textbf{95.51}   \\
                                                  \multicolumn{2}{l|}{DSSCC \cite{kumar2022intent} }            & 69.55&63.13&85.37 &  70.55& 56.89& 68.92 &  83.51&79.81&94.91    \\
                                                  \multicolumn{2}{l|}{DKT \cite{mou2022disentangled} }            & 70.38	&61.16	&83.71    &  72.57   & 58.6     & 69.12 &  75.44   & 71.20     & 92.69    \\
                                                  \multicolumn{2}{l|}{DPN \cite{an2023generalized} }            

& 70.23	&60.44	&\underline{85.7}   & 71.45	&62.61&	\textbf{77.86}  &82.76&	79.43&	\underline{95.4} \\
                                                  \multicolumn{2}{l|}{Llama2 \cite{touvron2023llama}}              &25.13	&	43.21&2.06    &69.26     &66.00      & 40.64&26.49&1.79&52.21\\
                                                    \hline
                                                  
                                                  \multicolumn{2}{l|}{PLPCL} &\textbf{76.50} &\textbf{67.13}	&\textbf{85.99}	& 	\textbf{77.63} & \textbf{63.58}     & 72.2 & 	\underline{86.38} & \underline{81.65}     & 95.02      
                                                       \\ \hline

\end{tabular}
}
\caption{The open-setting results on three datasets. Overall 1$^{st}$/2$^{nd}$ in \textbf{bold}/{\underline{underline}}. We randomly sample 30\% of all classes as OOD intents for both datasets. Results are averaged over three random runs. (p $<$ 0.05 under t-test) 
}
\label{tab:main_result2}
\vspace{-0.3cm}
\end{table*}

\subsection{Baselines} 
We employ a series of semi-supervised methods as baselines for comparing OOD intent discovery and open intent discovery: DTC\_BERT \cite{hsu2017learning}, KCL\_BERT \cite{han2019learning}, MCL\_BERT \cite{hsu2019multi}, CDAC+ \cite{lin2020discovering}, DeepAligned \cite{zhang2021discovering}, DSSCC \cite{kumar2022intent}, DKT \cite{mou2022disentangled},DPN \cite{an2023generalized}. We also evaluate the classification ability of large language models with Llama2 \cite{touvron2023llama}. To ensure lightweight implementation and reduce reliance on external data, we excluded MTP-CLNN \cite{zhang2022new}, as it requires extensive use of externally labeled data during the pre-training phase. All baselines use the same BERT backbone to ensure a fair comparison. 

\subsection{Evaluation Matrics}
We use three cluster evaluation metrics ACC, ARI, and NMI to evaluate the model effect, followed by DeepAligned \cite{zhang2021discovering}.  To obtain the results of ACC, we use the Hungarian algorithm to map prediction categories to ground-truth.

\subsection{Implementation Details}
We use the pre-trained bert-base-uncased model as the backbone consistent with the previous work, 
and pooling the context embeddings for each token using GRU and dense layers. The feature vector dimension is 768, the dropout probability is 0.1, and the GRU layer number is 1. In OOD discovery, the batch size of IND pre-training is 128, in the OOD clustering stage, the batch size of STACKOVERFLOW-OOD and BANKING-OOD are both 400, and the batch size of CLINC-OOD is 512. For open intent discovery, the batch\_size is 128 for each dataset for the pre-training and training stages. As with DKT, the learning rate of the pre-training process is set to 5e-5 of the training process is set to 0.0003, and the instance-level feature dimension is 128. Therefore, the cluster prototype feature dimension is also 128. 
The training epochs for training stage are 100. The experiment was conducted on an RTX 2080Ti GPU, and the running process takes 4 hours. We set $k=2$ for pseudo-label threshold selection and we also analyze the performance of different confidence thresholds. We reproduce DSSCC \cite{kumar2022intent} and DPN \cite{an2023generalized} under the settings outlined in the original papers. For BANKING-OOD and CLINC-OOD, the results of CDAC+, DeepAligned and DKT are obtained from \cite{mou2022disentangled}, and others are obtained from the text open intent recognition platform \cite{zhang2021textoir}. For open-setting, the results of baselines except DKT, DSSCC and DPN  are obtained from \cite{zhang2021textoir}.

\subsection{Main Results}
Table \ref{tab:main_result1} and \ref{tab:main_result2} show the performance of different models under the two task settings of three datasets. Our method outperformed previous approaches, especially on the BANKING and STACKOVERFLOW datasets, 
indicating its strong adaptability to single-domain intent classification tasks and superior discriminability for professional intents with semantically similar meanings. For the CLINC dataset, which is a cross-domain dataset with significant differentiation between categories, previous methods also exhibit strong clustering capabilities.

\subsection{Comparison with Large Language Model}
In the penultimate row of the experiment results Table \ref{tab:main_result1} and \ref{tab:main_result2}, we compared our results with Llama2-13B model \cite{touvron2023llama}. Taking into account the input tokens' limitations and their relatively weaker clustering abilities, we employed large models for classification tasks with the provision of category names. 
The example of a prompt template is shown in Appendix \ref{sec:template}.
The results indicate that the performance of LLM is inferior to our method in both settings of the three datasets. On the STACKOVERFLOW dataset with few categories, LLM outperforms some previous methods. However, on the BANKING and CLINC datasets with a larger number of categories, LLM clustering shows poor performance.

\subsection{Ablation Study and Further Analysis}
\begin{table}[ht]
\centering
\small
\resizebox{0.4\textwidth}{!}{
\begin{tabular}{l|c|c|c}
\hline
               & ACC   & ARI   & NMI    \\ \hline
ILCL,CLCL & 46.07 & 36.79 & 37.00 \\
\ \ \ \ +SCL,CE          & 70.38 & 61.16 & 83.71 \\ 
\ \ \ \ \ \ \ \ \ \ +PCL            & 74.29 & 65.90 & 85.36  \\ \hline
\ \ \ \ \ \ \ \ \ \ \ \ \ +PL         & \textbf{76.50} & \textbf{67.13} & \textbf{85.99} \\ \hline
\end{tabular}}
\vspace{-0.2cm}
\caption{Ablation study of different training objectives on BANKING-open.}
\label{tab:ablation study}
\vspace{-0.3cm}
\end{table}

\begin{table}[t]
\centering
\small
\resizebox{0.35\textwidth}{!}{
\begin{tabular}{l|c|c|c}
\hline
             $\sigma$   & ACC   & ARI   & NMI    \\ \hline
1 &  74.29 & 65.90 & 85.36 \\
0.99        &\textbf{76.50} & \textbf{67.13} & 85.99 \\ 
0.9         & 76.43 & 66.25 & \textbf{86.16}  \\
0.8         & 74.38 & 65.25 & 85.48 \\
0.7         &73.18 & 64.75 & 85.35 \\
0.5         &70.78 & 62.39 & 84.19 \\ 
0           &66.17 & 17.66 & 80.72 \\ \hline
\end{tabular}}
\vspace{-0.2cm}
\caption{Results under different confidence thresholds on BANKING-open.}
\label{tab:threshold}
\vspace{-0.6cm}
\end{table}

Table \ref{tab:ablation study} shows the effects of each module in our model, experimenting on BANKING-open. The results show that including SCL and CE during training helps to fully utilize the supervised signal. The absence of the supervised signal will result in a partial loss of pre-training information and a significant decrease in effectiveness. The addition of PCL improves the model's performance by 3.91\% (ACC), 4.74\% (ARI), and 1.65\% (NMI), indicating that explicitly decoupling and separating the cluster centers is beneficial for distinguishing and separating different category features in the feature space.
The addition of confident pseudo-labels (PL) improved the model's performance by 2.21\% (ACC), 1.23\% (ARI), and 0.63\% (NMI), indicating that gradually including samples with sufficiently high confidence in the supervised signal during model iteration is beneficial for obtaining prior information, compensating for the limitations of simple sample scattering in unsupervised contrastive learning.

Table \ref{tab:threshold} shows the effect of different confidence thresholds on the effectiveness of the model on BANKING-open. When the threshold is 1, no pseudo-labels are used. When the threshold is 0, pseudo-labels are used for all samples. The confidence threshold analysis of on other datasets and settings can be found in Appendix \ref{sec:loss weights}.

Figure \ref{Effect of labeled ratio} illustrates the clustering performance (ACC)  of various models at different labeled ratios for IND intents while maintaining an IND category ratio of 70\%. The results demonstrate the robust performance of our models across different labeling ratios. More results showing the effects of labeled ratio and known cluster ratio can be checked in Appendix \ref{sec:loss weights}.
 

\begin{figure}[h]
    \centering
    \centering
  \vspace{-0.3cm}
        \includegraphics[scale=0.3]{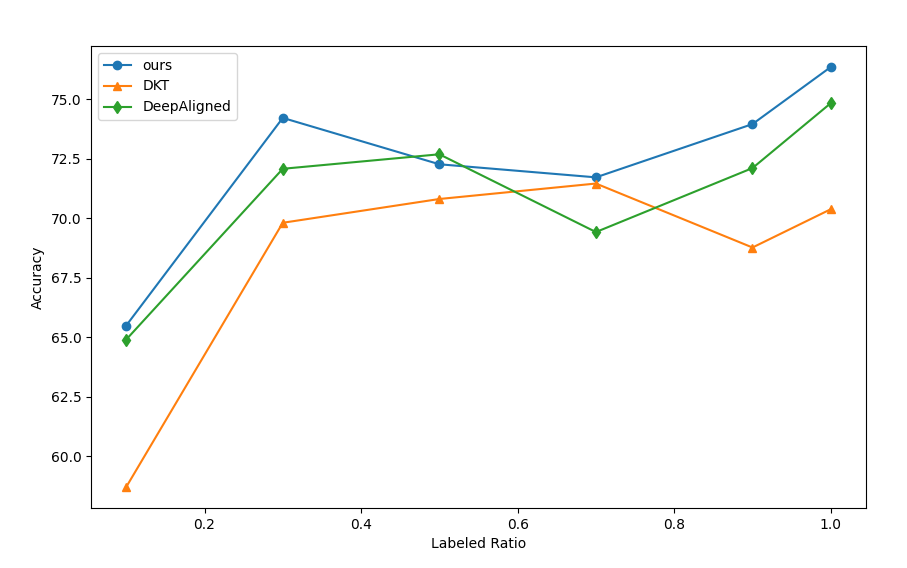}

    \vspace{-0.3cm}
    \caption{ Influence of the labeled ratio on BANKING-open. 
    }
    \label{Effect of labeled ratio}
    
    \vspace{-0.5cm}
\end{figure}

\subsection{\textit{K}-estimate Experiments} 

\begin{table*}[t]
\centering

\resizebox{1\textwidth}{!}{%
\begin{tabular}{ll|lll|lll|lll}
\hline

\multicolumn{2}{c|}{\multirow{2}{*}{Method}}                                          & \multicolumn{3}{c|}{BANKING-Open} & \multicolumn{3}{c|}{Stackoverflow-Open} &\multicolumn{3}{c}{CLINC-Open}  \\
                                                         &                & ACC      & ARI      & NMI      & ACC      & ARI      & NMI    & ACC  & ARI     & NMI       \\ \hline

                                                  \multicolumn{2}{l|}{DeepAligned \cite{zhang2021discovering}
                                                  }                      &  66.79&	55.58&	82.17      &  60.48	&42.09	&67.04& 75.96&	72.92&	93.33       \\

                                                  \multicolumn{2}{l|}{DPN \cite{an2023generalized} 
                                                  }            &69.77	&62.22&	\textbf{86.38}   &65.6	&45.91&	66.85 &78.36	&\textbf{75.14}&	\textbf{94.53}\\
                                                  
                                                  \multicolumn{2}{l|}{Ours} &  \textbf{71.17}&	\textbf{63.58}	&84.74&
                                                  \textbf{69.27}&	\textbf{55.96}	&\textbf{68.30}&\textbf{80.58}	&74.56&	94.16      
                                                       \\ \hline

\end{tabular}
}
\caption{Clustering performance with predicted \textit{K} 
}
\label{tab:k_results}
\vspace{-0.3cm}
\end{table*}
In real-world scenarios, it is difficult to determine the number of clusters \textit{K} in advance. Therefore, we conduct \textit{K}-estimation experiments by evaluating both the accuracy of the predicted \textit{K} value and the performance using the predicted \textit{K} following the approach in DeepAligned \cite{zhang2021discovering}.


We compared the performance of our method with the DeepAligned and DPN \cite{an2023generalized} algorithms and show the results in Table \ref{tab:k}. Our predicted \textit{K} values are closer to the ground truth compared to those of DeepAligned and DPN, demonstrating that the features extracted by our PLPCL model are more clustering-friendly, resulting in a more accurate estimation of the number of intents.
\begin{table}[h]
\centering
\small
\resizebox{0.5\textwidth}{!}{
\begin{tabular}{l|c|c|c}
\hline
             \textit{K}   & BANKING   & Stackoverflow   & CLINC    \\ \hline
Ground Truth &  77&	20&	150 \\
DeepAligned       &66 & 15 & 130 \\ 
DPN       & 67&	18&	137 \\
Ours       & 68	&19	&138 \\  \hline
\end{tabular}}
\caption{ Estimation of the number of categories.}
\label{tab:k}
\end{table}
Based on our predicted \textit{K}, we conducted experiments on three datasets to evaluate the impact of discrepancies between the estimated and actual number of unknown intents on clustering performance. The experimental results presented in Table \ref{tab:k_results} indicate that our approach outperforms the baseline methods when the number of clusters is unknown.


\section{Conclusion}
In this paper, we propose a pseudo-label enhanced prototypical contrastive learning approach for both open intent discovery and OOD intent discovery. 
The pseudo-label filtering strategy enhances supervised signal during the training process, while the prototypical contrastive learning module addresses the isolation issue between two independent contrastive learning heads.
Experiments on two task settings and three benchmark datasets demonstrate the effectiveness of our proposed method.
We hope to explore more self-supervised methods for OOD and open intent discovery in the future.

\section*{Limitations} 

In this work, we employ BERT-style models as our backbone, while the disadvantage lies in lacking the ability to generate coherent and contextually relevant text. While they can handle known intents well, BERT-style models may struggle with completely novel or OOD intents as they rely heavily on patterns seen during training. The generative LLMs are excellent at generating human-like text, making them suitable for creating responses and discovering new intents in open-ended queries. These models can adapt to new or unseen intents more effectively by generating diverse responses based on the input context, which is valuable for open intent discovery. To leverage the strengths of both BERT-style models and generative LLMs, some hybrid approaches can be employed in the future.

\section*{Acknowledgments}
This work is in part funded by the NSFC, China under Grants 62372364; in part by China Postdoctoral Science Foundation (2020M683496); in part by Shaanxi Provincial Technical Innovation Guidance Plan, China under Grant 2024QCY-KXJ-199; in part by Key R\&D Plan of Xianyang City, China under Grant L2023-ZDYF-QYCX-002; and in part by the National Postdoctoral Innovative Talents Support Program, China (BX20190273).

\bibliography{custom}

\nocite{zeng2021modeling}
\nocite{wu-etal-2022-revisit}
\nocite{he-etal-2020-contrastive}
\nocite{DBLP:journals/corr/abs-2110-13691}
\nocite{an2022finegrained}
\nocite{devlin2019bert}
\nocite{larson-etal-2019-evaluation}
\nocite{casanueva-etal-2020-efficient}
\nocite{chatterjee-sengupta-2020-intent}
\nocite{xu-etal-2015-short}
\nocite{kuhn1955hungarian}
\nocite{liu2022residual}
\nocite{de2019continual}
\nocite{wuenhancing}
\nocite{mou2023bridge}
\nocite{song2023large}
\nocite{zhou2023towards}
\nocite{van2023camell}
\nocite{an2023generalized1}
\nocite{chen2022co}
\nocite{lin2024flippedclassroom}
\appendix

\section{Appendix}
\label{sec:appendix}
\subsection{Further Analysis of loss weights and settings}
\label{sec:loss weights}
\begin{figure}[h]
    \vspace{-0.5cm}
    \centering
    \centering
   \includegraphics[height=0.35\textwidth]{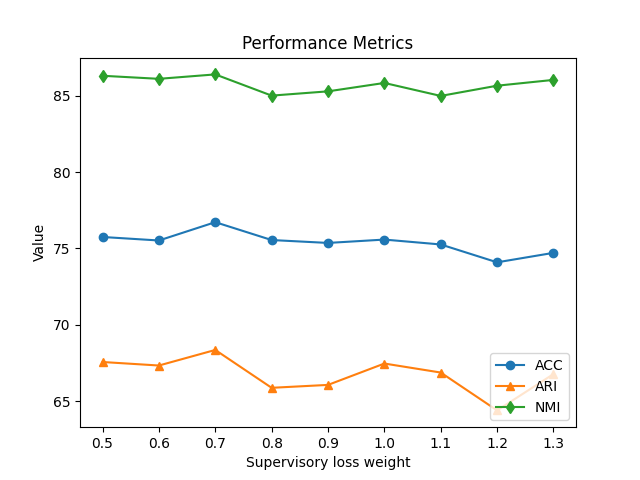}
    \caption{Influence of the supervisory loss weight on BANKING dataset
    }
    \label{loss_weight}
    \vspace{0.5cm}
\end{figure}
\begin{figure*}[h]
    \centering
    \centering
    \subfigure{
        \includegraphics[scale=0.5]{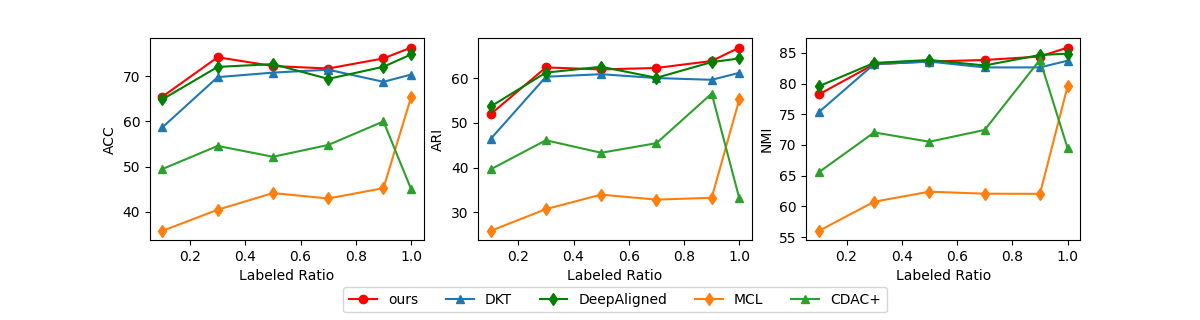}
    }
    \subfigure{
        \includegraphics[scale=0.5]{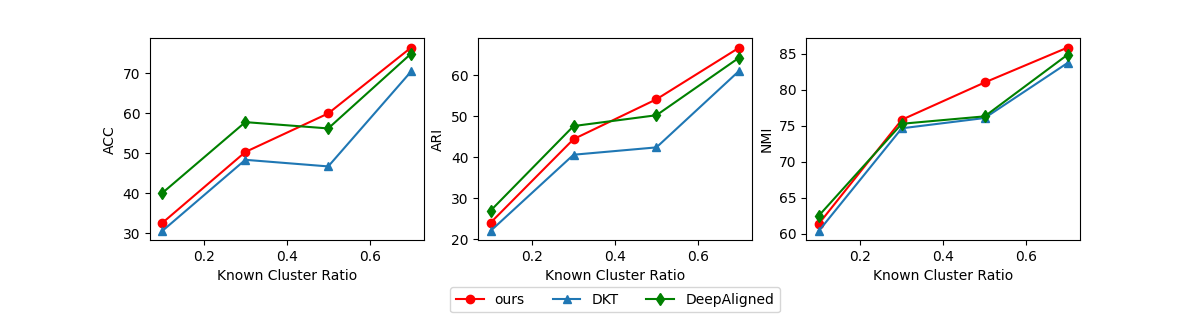}
    }
    \vspace{-0.5cm}
    \caption{ Influence of the labeled ratio and known cluster ratio on BANKING-open. 
    }
    \label{Effect of labeled ratio and known cluster ratio}
    
    \vspace{-0.5cm}
\end{figure*}

Figure \ref{loss_weight} demonstrates the performance of our model on the BANKING dataset under different weights of supervised contrastive loss, showing that our model is insensitive to loss weights. Figure \ref{Effect of labeled ratio and known cluster ratio} illustrates the impact of different labeled ratios and known cluster ratios on the model performance.

\begin{table}[h]
\centering
\small
\resizebox{0.3\textwidth}{!}{
\begin{tabular}{l|c|c|c}
\hline
             $\sigma$   & ACC   & ARI   & NMI    \\ \hline
1 &  74.33	&60.66	&70.92 \\
0.99        &\textbf{77.63} & \textbf{63.58} & 72.2 \\ 
0.9        & 71.08	&57.74&	68.38  \\
0.8        & 69.37&	56.76&	68.8 \\
0.7        & 66.93&	54.24&	67.64 \\
0.5       &  66.9	&57.15&	70.18 \\ 
0          & 66.8	&37.73&	\textbf{72.7} \\ \hline
\end{tabular}}
\vspace{-0.2cm}
\caption{Results under different confidence thresholds on STACKOVERFLOW-open.}
\label{tab:threshold_stackoverflow_open}
\vspace{-0.1cm}
\end{table}
\begin{table}[h]
\centering
\small
\resizebox{0.3\textwidth}{!}{
\begin{tabular}{l|c|c|c}
\hline
             $\sigma$   & ACC   & ARI   & NMI    \\ \hline
1 &  67.52	&\textbf{55.05}&	69.27 \\
0.99        &\textbf{68.37} & 53.19 & \textbf{72.04} \\ 

0.9        & 63.09	&49.68&	70.19 \\
0.8        & 53.26	&38.04&	62.4 \\
0.7       &  47.93&	36.88	&63.02 \\ 
0.5       &  34.24	&21.7&	49.64 \\  \hline
\end{tabular}}
\vspace{-0.2cm}
\caption{Results under different confidence thresholds on BANKING-OOD.}
\label{tab:threshold_banking_OOD}
\vspace{-0.1cm}
\end{table}
\begin{table}[h]
\centering
\small
\resizebox{0.3\textwidth}{!}{
\begin{tabular}{l|c|c|c}
\hline
             $\sigma$   & ACC   & ARI   & NMI    \\ \hline
1 &  85.17&	67.58&	65.73 \\
0.99        &\textbf{86.28} & \textbf{69.64} & \textbf{66.95} \\ 

0.9        & 71&	53.67&	56.12 \\
0.8        & 71.67	&43.63	&49.54 \\
0.7       &  53.39	&25.06	&41.03 \\ 
0.5       &  61.72	&43.61	&51.77 \\  \hline
\end{tabular}}
\vspace{-0.2cm}
\caption{Results under different confidence thresholds on STACKOVERFLOW-OOD.}
\label{tab:threshold_stakoverflow_OOD}
\vspace{-0.2cm}
\end{table}

Table \ref{tab:threshold_stackoverflow_open} ,\ref{tab:threshold_banking_OOD}, \ref{tab:threshold_stakoverflow_OOD} shows the impact of different confidence thresholds on the model's effectiveness across various datasets and task settings. The experiments show that selecting an appropriate confidence threshold is crucial for optimizing model performance across different datasets and tasks. Higher confidence thresholds generally lead to better classification results, indicating that the model can more accurately identify and leverage high-confidence samples for effective learning. This effect is even more pronounced in the OOD setting, where a low threshold can significantly degrade performance, emphasizing the need for careful threshold selection.

\subsection{Visualization}

\begin{figure*}[h]
    \centering
    \centering
    \subfigure[DKT]{
        \includegraphics[scale=0.4]{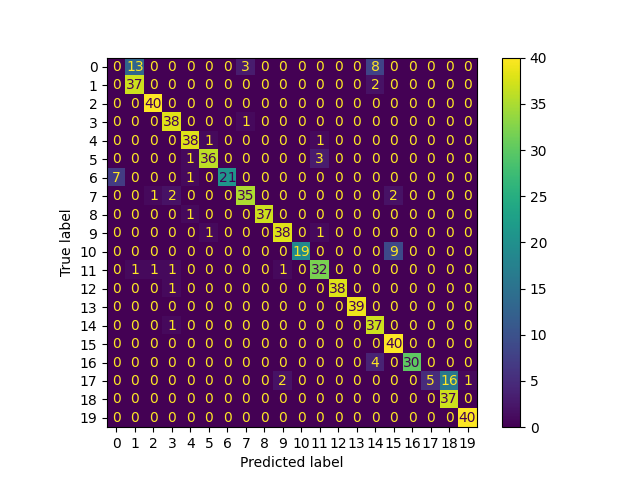}
    }
    \subfigure[Ours]{
        \includegraphics[scale=0.4]{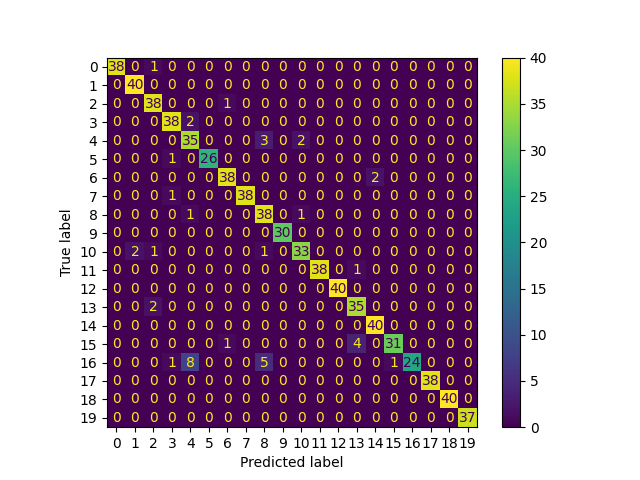}
    }
   
    \vspace{-0.5cm}
    \caption{Confusion matrix visualization of different models. We use the same test set of BANKING-open. 
    }
    \label{confusion matrix visual}
    \vspace{-0.5cm}
\end{figure*}

\begin{figure*}[h]
    \centering
    \centering
    \subfigure[DKT\_OOD]{
        \includegraphics[scale=0.25]{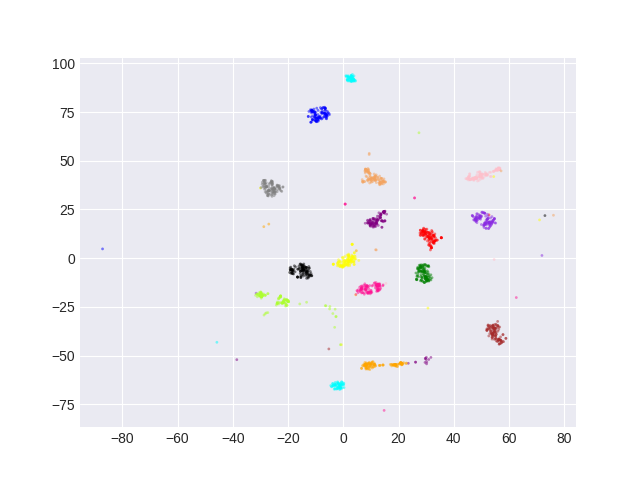}
    }
    \subfigure[Ours\_OOD]{
        \includegraphics[scale=0.25]{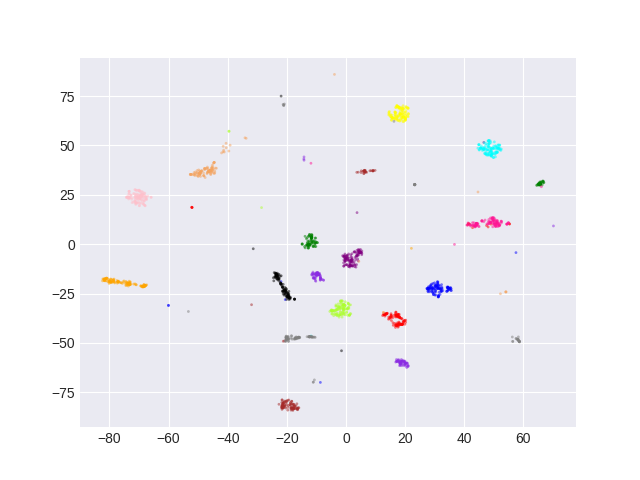}
    }
    \\
    \subfigure[DKT\_open]{
        \includegraphics[scale=0.25]{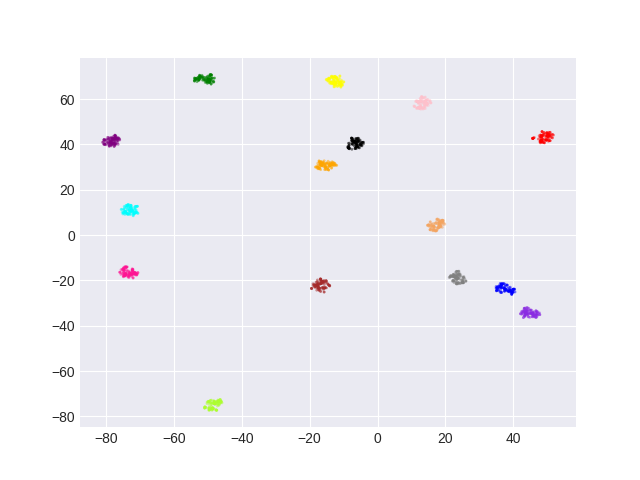}
    }
    \subfigure[Ours\_open]{
        \includegraphics[scale=0.25]{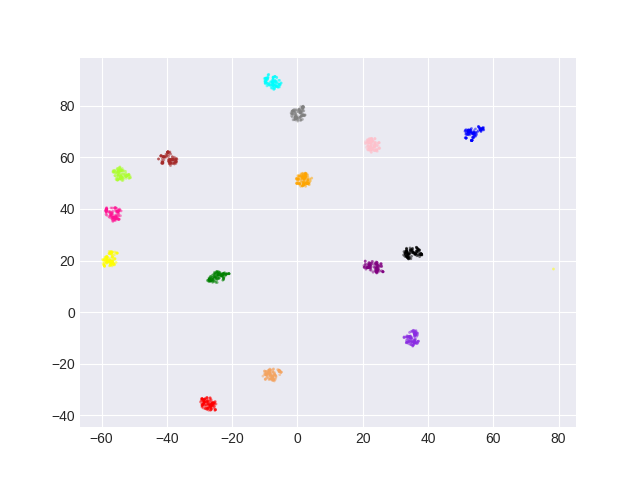}
    }
    \vspace{-0.5cm}
    \caption{Cluster visualization of different models. We use the same test set of BANKING.}
    \label{cluster visual}
    \vspace{-0.5cm}
\end{figure*}

Figure \ref{confusion matrix visual} shows the visualization results of the confusion matrix for DKT and our model on BANKING-open, with a total of 77 categories in the test set, and we show the first 20 categories. It can be found that the DKT model may completely confuse two certain categories, i.e. the samples of two certain  categories are grouped into the same cluster. However, our model avoids this problem well, and rarely there is no correct sample in a certain category. 
Figure \ref{cluster visual} shows the clustering visualization results of DKT and our model on BANKING-open and BANKING-OOD. For fair comparison, we use the same representation after the pooling layer. We can find that after adding contrastive learning for prototype and reliable pseudo label, while keeping the samples of the same cluster compact, the distance between different clusters is widened, and the different clusters become scattered on the whole feature space.Unlike scattering of unlabeled samples, the premise for contrastive learning in cluster prototype is that each cluster has its own unique features, and cluster center scattering aims to separate these features.

\subsection{Prompt Template of LLM}
\label{sec:template}
The prompt template is shown in Table\ref{tab:template}.
\begin{table}[h]
\centering
\small
\resizebox{0.5\textwidth}{!}{
\begin{tabular}{l  }
\hline

\hline
             Below is an instruction that describe a task. Write a response 
             \\that appropriately completes the request.     \\ 
             $\#\#\#$instruction: \\ 
             Please give the intent label for the following sentences.\\
             Select one label in the set \{...\}\\
              For example:\\
                Input:\\
              Can I exchange currencies with this app? .\\
                Output:\\
               \{  intent\_label:"exchange\_via\_app"\}\\
                $\#\#\#$question:\\
               Input:\\
             \{data sample\}\\
             Provide intent label in JSON format with the following keys: intent\_label\\
             $\#\#\#$Response: \\
\hline

\hline
\end{tabular}}
\caption{ An example of the prompt templates we used.}
\label{tab:template}
\end{table}


\end{document}